\documentclass[conference,compsoc]{IEEEtran}
\ifCLASSOPTIONcompsoc
  \usepackage[nocompress]{cite}
\else
  \usepackage{cite}
\fi
\ifCLASSINFOpdf
  \usepackage[pdftex]{graphicx}
\else
\fi
\usepackage{amsmath}
\begin{document}
%
\title{Soft Sensing Transformer:\\ Hundreds of Sensors are Worth a Single Word}

\makeatletter
\newcommand{\linebreakand}{%
  \end{@IEEEauthorhalign}
  \hfill\mbox{}\par
  \mbox{}\hfill\begin{@IEEEauthorhalign}
}
\makeatother
\author{
  \IEEEauthorblockN{Chao Zhang}
  \IEEEauthorblockA{\textit{Seagate Technology, MN, US} \\
    \textit{University of Chicago, IL, US}\\
    chao.1.zhang@seagate.com}
  \and
  \IEEEauthorblockN{Jaswanth Yella}
  \IEEEauthorblockA{\textit{Seagate Technology, MN, US} \\
    \textit{University of Cincinnati, OH, US}\\
     jaswanth.k.yella@seagate.com}
  \and
  \IEEEauthorblockN{Yu Huang}
  \IEEEauthorblockA{\textit{Seagate Technology, MN, US} \\
    \textit{Florida Atlantic University, FL, US}\\
    yu.1.huang@seagate.com}
  \linebreakand 
  \IEEEauthorblockN{Xiaoye Qian}
  \IEEEauthorblockA{\textit{Seagate Technology, MN, US} \\
    \textit{Case Western Reserve University, OH, US}\\
    xiaoye.qian@seagate.com}
  \and
  \IEEEauthorblockN{Sergei Petrov}
  \IEEEauthorblockA{\textit{Seagate Technology, MN, US} \\
    \textit{Stanford University, CA, US}\\
    sergei.petrov@seagate.com}
  \and
  \IEEEauthorblockN{Andrey Rzhetsky}
  \IEEEauthorblockA{\textit{University of Chicago} \\
    \textit{IL, US}\\
    arzhetsky@medicine.bsd.uchicago.edu}
  \linebreakand 
  \IEEEauthorblockN{Sthitie Bom}
  \IEEEauthorblockA{\textit{Seagate Technology} \\
    \textit{MN, US}\\
    sthitie.e.bom@seagate.com}
}


%


\IEEEoverridecommandlockouts
\IEEEpubid{\makebox[\columnwidth]{This paper has been accepted by 2021 IEEE International Conference on Big Data \hfill} \hspace{\columnsep}\makebox[\columnwidth]{ }}

\maketitle
\IEEEpubidadjcol
\begin{abstract}
With the rapid development of AI technology in recent years, there have been many studies with deep learning models in soft sensing area. However, the models have become more complex, yet, the data sets remain limited: researchers are fitting million-parameter models with hundreds of data samples, which is insufficient to exercise the effectiveness of their models and thus often fail to perform when implemented in industrial applications. To solve this long-lasting problem, we are providing large scale, high dimensional time series manufacturing sensor data from Seagate Technology to the public. We demonstrate the challenges and effectiveness of modeling industrial big data by a Soft Sensing Transformer model on these data sets. Transformer is used because, it has outperformed state-of-the-art techniques in Natural Language Processing, and since then has also performed well in the direct application to computer vision without introduction of image-specific inductive biases.  We observe the similarity of a sentence structure to the sensor readings and process the multi-variable sensor readings in a time series in a similar manner of sentences in natural language. The high-dimensional time-series data is formatted into the same shape of embedded sentences and fed into the transformer model. The results show that transformer model outperforms the benchmark models in soft sensing field based on auto-encoder and long short-term memory (LSTM) models. To the best of our knowledge, we are the first team in academia or industry to benchmark the performance of original transformer model with large-scale numerical soft sensing data. Additionally, In contrast to the natural language processing or computer vision tasks where human-level performances are regarded as golden standards, our large scale soft sensing study is an example that transformer goes beyond human, because the high dimensional numerical data is not interpretable for human.
\end{abstract}


%
\IEEEpeerreviewmaketitle

\section{Introduction}
In the last decades, the development of smart sensors has attracted a lot of attention from government, academia and industry.
The European Union’s 20-20-20 goals (20\% increase in
energy efficiency, 20\% reduction of CO2 emissions, and 20\%
renewable by 2020) rely on smart metering as one of their key
enablers. Smart meters usually involve real-time or near real-time
sensors, notification and monitoring. In 2013, Germany
proposed the concept of Industry 4.0, the main aim of which is
to develop smart factories for producing smart products. The
US government in September 2020 announced that the US is
providing more than \$1 billion towards establishing research
and hubs for Industry 4.0 technologies. Singapore's current
five-year SU\$13.8 billion R\&D is injecting more funds into
expanding fields such as advanced manufacturing. China's
China Manufacturing 2025 goal is also to make the
manufacturing process more intelligent. These initiatives
require that we have better sensing technologies to understand
and drive our processes. Sensors have the potential to contain
information about process variables which can be exploited by
data-driven techniques for smarter monitoring and control of
manufacturing processes. Soft sensing is the general term used
for the approaches and the algorithms that are used to estimate
or predict certain physical quantities or product quality in the
industrial processes based on the available sensing modalities,
measurements, and knowledge.

As the industrial process have become more complicated and the size of available data has increased dramatically, there has been growing body of research on deep learning methods with applications in the soft sensing field. A recent survey on deep learning methods for soft sensor \cite{sun2021survey-softsensing} has illustrated the significance of the deep learning applications and reviewed the most recent studies in this field. The deep learning models are mostly based on autoencoder \cite{liou2014autoencoder}, restricted Boltzmann machine \cite{smolensky1986information}, convolutional neural network \cite{lecun2015deep}, and recurrent neural network \cite{rumelhart1986learning}. The applications varies from traditional factories \cite{sqae} to wearable IoT devices \cite{qian2019smart, qian2020wearable}

There has been a variety of novel deep learning models such as variational autoencoder models which attempts to enhance the representation ability or augment the data \cite{huang2020reliable, huang2021prognostics}, semi-supervised ensemble learning model that quantifies the contribution of different hidden layers in stacked autoencoder \cite{sun2020deep}, and gated convolutional transformer neural network that combines several state-of-art algorithms together to deals with a time-series data set \cite{geng2021novel}. As the deep learning models become more and more complex, their capabilities to handle complex processes and large data sets also increase. However, in these  studies researchers are still using very small data sets such as wastewater treatment plant and Debutanizer column \cite{Souza2013_jpc, fortuna2007debutanizer} containing low dimensional data with only hundreds to thousands of data samples. These small data sets are not sufficient to illustrate the effectiveness of these advanced deep learning models with millions of parameters. To solve this issue, we collected gigabytes of numerical sensor data from Seagate's wafer manufacturing factories in USA and Ireland. These data sets contain high-dimensional time-series sensor data that is collected directly from the Seagate wafer factories with only the necessary anonymization, and they are big, complex, noisy and impossible to interpret in their raw form by humans.. In this article, We evaluate a soft sensing transformer model against the most commonly methods applied to soft sensing problems including models based on autoencoder and LSTM \cite{hochreiter1997long}. The key components of the original transformer model is maintained and the other parts of the architecture are modified to fit into our data sets and tasks. 

Transformer, since it's proposal in 2017 \cite{vaswani2017attention}, together with it's derivatives such as BERT\cite{devlin2018bert}, have been the most active research topic in the natural language processing (NLP) field as well as the top performer in many NLP tasks\cite{lin2021survey-transformer}. Due to its extraordinary representative capability, transformer model has also shown equally good performance in the computer vision area \cite{han2020survey-vit}. First proposed in 2020, vision transformer \cite{dosovitskiy2020image-vit} and its variants have achieved the state-of-art performances on many computer vision benchmarks such as image classification, semantic segmentation and  object detection \cite{liu2021swin, zhai2021scaling}.

From texts in NLP, which can be regarded as categorical data, to images (two dimensional integer values) in computer vision, a natural further extension would be soft sensing data which is time series with continuous floating numbers. While the Bayes error rate \cite{fukunada1990introduction} in NLP and computer vision tasks are usually defined as human-level performance, our soft sensing task is impossible for a human to classify based on the hundreds of sensor values. We show in this paper that Transformer architecture not only works great for natural language and images, but also for numerical data, and it is able to represent the data that is not interpretable by human.

The rest of this paper is organized as the following: We discuss the soft sensing transformer model in 
section. \ref{section: method}, several industrial soft sensing data sets in section. \ref{section: data}, the results of the soft sensing transformer on these data sets in section. \ref{section: results}, and discussions and conclusions in section. \ref{section: conclusion}.

\section{Methodology} \label{section: method}
While implementing the soft sensing model, we follow the original transformer architecture as closely as possible. The input module of the model is modified to fit the time-series sensor data, and the output module is modified for multi-task classification problems. This is the first study for benchmark results on these large scale sensor data sets with deep learning methods, also the first study for transformer model applied on large scale numerical sensor data.

\begin{figure}[!t]
\centering
\includegraphics[width=0.9\linewidth]{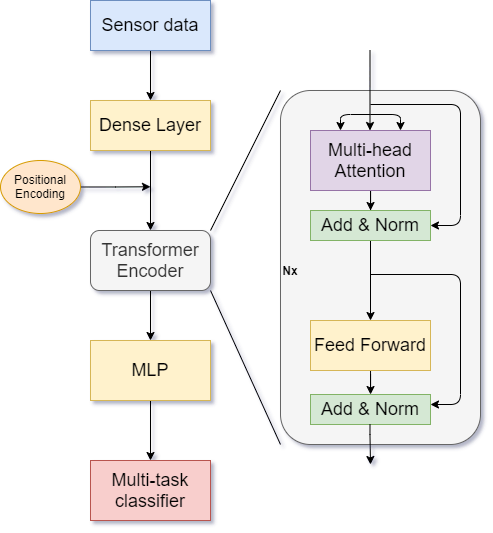}
\caption{Architecture of soft sensing transformer model}
\label{fig_sst}
\end{figure}

\subsection{Soft Sensing Transformer (SST)}
We illustrate the structure of the soft sensing transformer model in Fig. \ref{fig_sst}. Given that the data format of time-series sensor data is different from texts, we used a dense layer for the embedding at the starting point, which reduces the dimension of the input high dimensional sensor data. After this layer, the data format is the same as embedded sentences so that it can be feed into the transformer encoder without any modifications. Right before the encoder block, a positional encoding using sine and cosine functions of different frequencies is added as Equation. \ref{eqn:pe_even} and Equation. \ref{eqn:pe_odd} to cover the information of relative positions of different time steps. $PE$ stands for positional encoding, $pos$ is the position of a time step, and $dmodel$ is the dimension of embedded vectors. 

\begin{equation}
\label{eqn:pe_even}
PE_{(pos, 2i)} = sin(pos/10000^{2i/dmodel})
\end{equation}

\begin{equation}
\label{eqn:pe_odd}
PE_{(pos, 2i+1)} = cos(pos/10000^{2i/dmodel})
\end{equation}

Since the SST model requires a fixed size of input data, we added padding to samples with too few time steps so that each sample has the same time length. The time length is chosen as 99 percentiles of the sequence lengths in the raw data to cover most of the data and exclude outliers. The padding masks are also applied accordingly. In the encoder, multi-head scaled dot product attention, feed forward and residual connections are set up in the the same way as in the original transformer paper \cite{vaswani2017attention}. The multi-head attention is described as in Equation. \ref{eqn:mha}, the query, key and value are projected to $h$ heads with the weight matrices $W_i^q,W_i^k,W_i^v$. Each head has a dimension of $dmodel/h$, and a scaled dot-product attention is calculated for each head. Then the heads are concatenated and projected back to the original shape. 

\begin{scriptsize}
\begin{equation}
\label{eqn:mha}
MHA(Q, K, V) = Concat(softmax(\frac{QW_i^q(KW_i^k)^T}{\sqrt{d_k}})VW_i^v)W^o
\end{equation}
\end{scriptsize}

The Seagate data sets are contain measurement pass/fail information, and the SST model is built as an classification model. After the encoder blocks, a multi-layer perceptron (MLP) classifier is attached on top after a global average pooling. Because of the intrinsic complexity of the data, the classifier comprises a few individual binary classifiers. These binary classifiers partly share the input data and may be correlated with each other, resulting an inter-correlated multi-task problem (further discussed in Section. \ref{section: data}). In order to achieve the best performance in the multi-task learning, a weighting method based on uncertainty \cite{kendall2018multi-task} is applied, and we define the combined loss function as Equation. \ref{eqn:total_loss}:

\begin{equation}
\label{eqn:total_loss}
J  = \sum_i ( \frac{1}{\sigma_i^2}J_i + log\sigma_i)
\end{equation}

where $J$ is the total loss, $J_i$ is the loss of the $i_{th}$ classification task, and $\sigma_i$ is the uncertainty of the  $i_{th}$ classification loss, which is trainable during the model fitting.

\subsection{Optimization}
\subsubsection*{Data imbalance}
In the industrial settings, the data are highly imbalanced. As a classification model, we have only 1\% to 2\% of the data samples as positive. To deal with the imbalance, we experimented on both weighting methods and data sampling algorithms like SMOTE \cite{chawla2002smote}. We found that class weighting gives the best efficiency and performance in our experiments. The weight of the $j_{th}$ task, label $t$ ($0$ or $1$) is calculated based on the number of samples:

\begin{equation}
\label{eqn:weight}
w_j^t = \frac{N}{2m*n_j^t}
\end{equation}

in which $N$ is the total number of sample, $m$ is the number of tasks, and $n_j^t$ is the number of samples for label $t$ in the $j_{th}$ task. 

Combined with the uncertainty based multi-task learning as Equation. \ref{eqn:total_loss}, the final loss function of SST model is defined as weighted cross entropy:

\begin{equation}
\label{eqn:weighted_loss}
J = \sum_j^m\sum_{t=0}^1[\frac{1}{\sigma_{jt}^2} \sum_i^{n_j^t}  (y_{ijt}* log\hat{y}_{ijt}) + log(\sigma_{jt})]
\end{equation}

where $y_{ijt}$ and $\hat{y}_{ijt}$ are the true labels and predicted probabilities for the $i_{th}$ sample in task $j$ for label $t$. 
Note that the cross entropy loss is calculated in a multi-label classification manner and the loss for positive and negative cases are computed separately. We take $y_{ij1} = 1$ for positive samples, and $y_{ij0} = 1$ for negative samples. The weights for the positive and negative cases in a single binary classification task is also further tuned by $\sigma_{jt}$, which is the uncertainty or variance of the loss for label $t$ in task $j$. In this multi-task learning setting, we have $2m$ 'tasks' for the $m$ binary classifications. 

\subsubsection*{Activation functions}
For the transformer encoder part, a ReLU activation function \cite{nair2010rectified} is applied in the feed forward layer, which is consist of two dense layers that project the $dmodel$ dimensional vector to $dff$ dimension and project back to $dmodel$ dimension, respectively. ReLu activation function is set for the first dense layer in the feed forward layers. As for the MLP classifier, we applied sigmoid activation functions for all three layers because we found that it produced more stable results than ReLu in this case. 

\subsubsection*{Regularization} 
L2 regularizers are applied to all the dense layers in SST model, with a regularization factor of $10^{-4}$. Dropout \cite{srivastava2014dropout} is also applied to residual layers and embedding layers. We also applied dropout to each layer in the MLP block except for the final prediction layer. All dropout ratios are kept the same and a grid search in $[0.1, 0.3, 0.5]$ is performed to find the best dropout ratio.

\subsubsection*{Optimizer} 
We experimented with two kinds of optimizers: default adam optimizer \cite{kingma2014adam} with fixed learning rate, and scheduled adam optimizer similar as in \cite{vaswani2017attention}. The learning rate scheduled optimizer has shown a more stable result, so it's kept in further experiments.

For the scheduled adam optimizer, the parameters are set as $\beta_1 = 0.9$, $\beta_2 = 0.98$, $\epsilon=10^{-9}$. The learning rate is varied during the training process based on Equation. \ref{eqn:schedule}. $d$ is $dmodel$ in SST model, $step$ is the training step, and $warmup$ is set as $4000$. An extra $factor$ is added to tune the overall learning rate. A grid search for the $factor$ in $[0.1,0.3,0.5]$ is performed to find the optimal factor.
\begin{equation}
    \label{eqn:schedule}
    lr = factor*d^{-0.5}*min(step^{-0.5}, \frac{step}{warmup^{1.5}})
\end{equation}

\subsubsection*{Hyper-parameter tuning}
There are a few hyper-parameters to be tuned for the SST model training. As shown in Table. \ref{hyperparameters}, in total 7 hyper-parameters are tuned using a grid search. The hyper-parameters include number of the encoder block ($n\_layer$), the size of embedding layer ($dmodel$), the size of feed forward layer ($dff$), the dropout ratio, learning rate factor as in Equation. \ref{eqn:schedule}, batch size, number of heads for the multi-head attention layer ($n\_heads$), and whether or not to use the uncertainty based weighting as in Equation. \ref{eqn:total_loss}.
For the process of grid search, a smaller size of data are randomly sampled from the data sets, which contains 5000 samples for training and 3000 for validation. The best model is picked based on the validation results, evaluated by the area under a Receiver Operating Characteristic Curve (ROC-AUC) \cite{fawcett2006introduction}.

\begin{table}[!t]
\caption{Hyper-parameter search space}
\label{hyperparameters}
\centering
\begin{tabular}{cc}
\hline
Hyper-parameter & Values\\
\hline
$n\_layers$ & 2, 3, 4 \\
$dff$ & 32, 128, 512 \\
$dmodel$ & 64, 128, 256 \\ 
$dropout\_ratio$ & 0.1, 0.3, 0.5 \\
$learning\ rate\ factor$ & 0.1, 0.3, 0.5 \\
$batch\ size$ & 512, 1024, 2048 \\
$n\_heads$ & 1, 2, 4 \\
$uncertainty\ based\ weighting$ & on, off \\
\hline
\end{tabular}
\end{table}

\subsubsection*{Early-stopping}
Instead of setting a epoch number, we used an Keras early-stopping callback method in the training processing, with a patience of 100 epochs, and restore\_best\_weights=True. In this way, we got an optimized epoch number for each experiment without manually tuning.

\subsubsection*{Hardware} 
All the models are trained on an AWS instance with an NVIDIA Tesla V100 SXM2 GPU. It took around 20ms for each step, and about 30 minutes for the entire training. The grid search for hyper-parameters took about 36 hours. All the models are written with TensorFlow \cite{abadi2016tensorflow} version 2.2 and Keras\cite{chollet2015keras}. 

\section{Data} \label{section: data}

\begin{figure}[!t]
\centering
\includegraphics[width=0.9\linewidth]{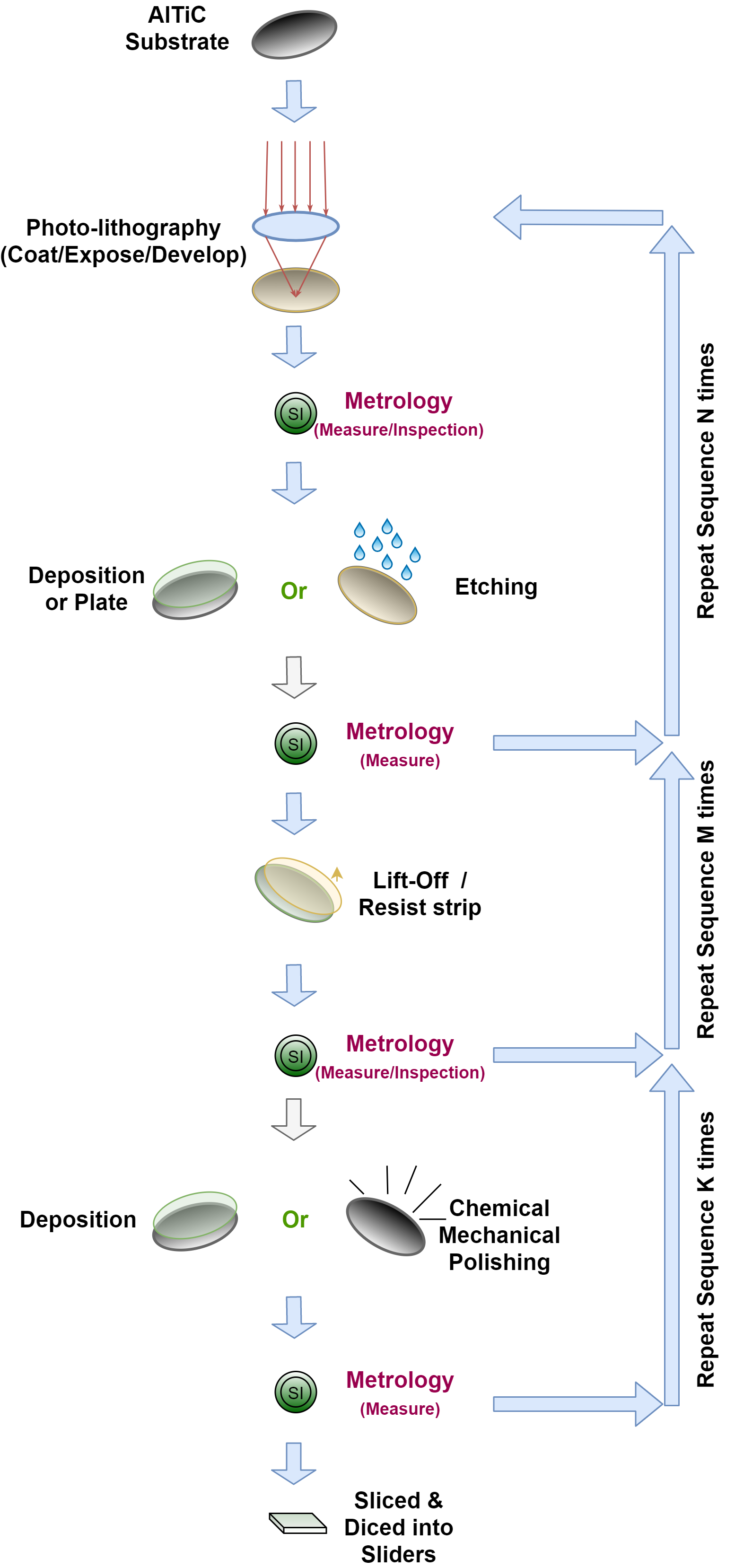}
\caption{High-level workflow of wafer manufacturing. Each wafer goes through multiple processing stages, each stage has corresponding metrology, in which a few quality control measurements are performed. The measurement results are used to decide whether the wafer is in a good shape to go to the next stage. Figure from https://github.com/Seagate/softsensing\_data.}
\label{fig_wafer}
\end{figure}

To fill the gap of publicly available large scale soft sensing data sets, we queried and processed several gigabytes of data sets from Seagate manufacturing factories in both the US and Ireland. These data sets contain high dimensional time-series sensor data coming from different manufacturing machines. 

As shown in Fig. \ref{fig_wafer}, to fabricate a slider used for hard drives, an AlTiC wafer goes through multiple processing stages including deposition, coating, lithography, etching, and polishing. Different products have different manufacturing lines, Fig. \ref{fig_wafer} shows a simplified and general processing. 
After each processing stage, the wafer is sent to metrology tools for quality control measurements. A metrology step may have a single or multiple different measurements made each of which could have varying degrees of importance.  

These processes are highly complex and are sensitive to both incoming as well as point of process effects. A significant amount of engineering and systems resources are employed to monitor and control the variability intrinsic to the factors that are known to affect a process. 

Metrology serves a critical function of managing these complexities for early learning cycles and quality control. This, however, comes at high capital costs, increased cycle time and considerable overhead to set up correct recipes for measurements, appropriate process control and workflow mechanisms. In each processing tool, there are dozens to hundreds of onboard sensors in the processing machines to monitor the state of the tool. These sensors collect information every few seconds and all these sensing values are collected and stored along with the measurement results.

\begin{figure}[!t]
\centering
\includegraphics[width=0.9\linewidth]{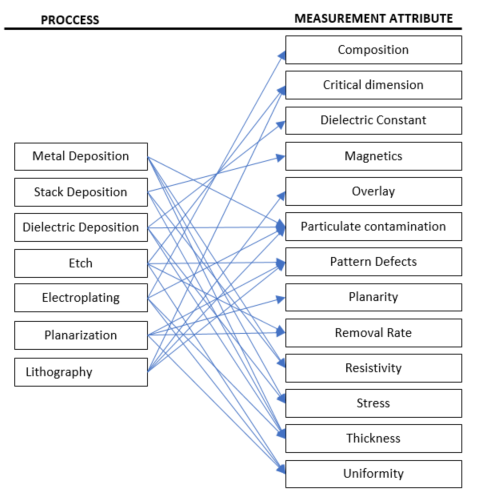}
\caption{Overview of the main categories of processes and the corresponding critical measurement variables per each category. Figure from https://github.com/Seagate/softsensing\_data.}
\label{fig_meas}
\end{figure}

As shown in Fig. \ref{fig_meas}, one time-series of sensor data are mapped to several measurements, and the same measurement can be applied to multiple processing sensor data points. Each measurement contains a few numerical values to indicate the condition of the wafers, and a decision of pass or fail is made based on these numbers. For the sake of simplicity, we only cover the pass/fail information for each measurement, so that each sample of time-series sensor data are mapped to several binary classification labels, resulting in a multi-task classification problem. On the other hand, some of measurements are linked to multiple processing stages, so that the SST model can learn the representations from one stage and apply to another stage when it's trained on data covering all the stages. Given this inter-correlation, training such a multi-task learning SST model leads to a better performance comparing with training the measurement tasks individually. From the perspective of industrial application, it's also more maintainable and scalable to have a single model instead of many ones.

The data sets in this paper cover 92 weeks of data. The first 70 weeks are taken as training data, and the following 14 weeks as validation data, last 8 weeks as testing data. The data sets are prepared by querying from raw data and doing some necessary pre-processing steps. While the sensors are collecting data every few seconds, there are a lot of redundancies, so we aggregated the data into short sequences. In each processing stage, a wafer goes through a few modules, and we aggregate the data by the module and get short time sequences. Other pre-processing steps on the data include a min-max scaling, imputation with neighbors, one-hot encoding for categorical variables, and necessary anonymization. The min-max scaler is fit only on training data, and applied on the entire data sets. Imputation is done by filling the missing values first by it's neighbors (a forward filling followed by a backward filling) if non-missing values exist in the same processing stage, otherwise filling by the mode of all the data. Categorical variables such as the processing stage information, the type of the wafer, the manufacturing machine in function, are one-hot encoded and concatenated to the sensor data as model input. As for the anonymization, only confidential information like the data headers is removed. 

Using data in different timezframes for training and testing reflects the application prospect of the SST model, because in this way the model can be directly deployed into factories once it performs well enough in testing data. However, this setting also makes it harder for the model to achieve a high performance because in reality there are too many uncontrollable factors in the factories and the data distribution of training and testing data may be different with each other.

These data sets are in Numpy format, which only include numerical values without any headers. Input files are rank 3 tensors with dimension (n\_sample, time\_step, features), and outputs are rank 2 tensors with dimension (n\_sample, 2*n\_tasks). Each binary classification task has two columns in the output file, first column for negative cases and second for positive cases. 

Three data sets are covered in this paper. They are from slightly different manufacturing tool families, and each has different processing stages and corresponding measurements. The number of samples for each measurement is summarized in Table. \ref{tbl:data}. More detailed information for each tool family is described below, and all the data are available at https://github.com/Seagate/softsensing\_data.

\begin{table}[!t]
\renewcommand{\arraystretch}{1.3}
\caption{Summary for the data sets: number of positive and negative samples for each task}
\label{tbl:data}
\centering
\begin{tabular}{ccccccc}
\hline
 & \multicolumn{2}{c}{P1} & \multicolumn{2}{c}{P2} & \multicolumn{2}{c}{P3} \\
Task &pos&neg&pos&neg&pos&neg\\
\hline
1&295&8328&256&6433&109&2496 \\
2&40&12747&773&26811&335&12857 \\
3&291&56198&2069&78844&46&1026 \\
4&188&14697&582&27809&15&4180 \\
5&568&40644&247&9652&300&22254 \\
6&863&84963&884&27337&166&40811 \\
7&2501&153970&2108&53921&875&75706 \\
8&490&2919&2016&77473&1097&18890 \\
9&104&29551&644&23305&537&4247 \\
10&57&10813&270&25651&1547&129914 \\
11&306&47219&3792&354328&& \\
\hline
\end{tabular}
\end{table}

\subsubsection*{P1}
 The sensor data are generated by a deposition tool that include both deposition and etching steps. There are 90 sensors installed in the tools and they capture data at a frequency of about every second. The critical parameters measured for this family of tools are magnetics, thickness, composition, resistivity, density, and roughness.
 
 After pre-processing mentioned above, there are 194k data samples in training, 34k samples in validation, and 27k samples in testing data. Each sample has 2 time steps, with 817 features. Some of the second time steps are missing and replaced with zero padding, and the 817 features come from 90 sensors, one-hot encoded categorical variables including the types of the wafer, the processing stages, and specific manufacturing tools etc, and a padding indicator as the last feature. 
 
 For the labels, there are 11 individual measurement tasks, each is a binary classification. We set the model output dimension as 22 to have separate predictions for negative and positive probabilities, and normalize them to get the predicted probabilities after applying class weights for the data imbalance. As shown in Table. \ref{tbl:data}, the data set are highly imbalanced, there are about 1.2\% of the samples have positive labels. 

\subsubsection*{P2} 
This second data set contains data generated by a family of ion milling (dry etch) equipment, which utilize ions in plasma to remove material from a surface of the wafer. There are 57 sensors for this data set, and the critical parameters measured for this family of tools are similar to P1 tools, but with slightly different measurement machines.

There are 457k training samples, 80k validation samples, and 66k testing samples in the data set. For this data set, there is no time-series information, but we treat it as 1 time step to fit into the same SST model. This data set is more complex in terms of categorical variables, resulting in 1484 features in total. 

The number of measurement tasks is 11, with an output dimension of 22, and about 1.9\% of the samples are positive as in Table. \ref{tbl:data}. Note that these 11 tasks are not the same as those in P1.

\subsubsection*{P3} 
The last data set is generated by sputter deposition equipment containing multiple deposition chambers, with unique targets. The number of sensors is 43, and critical parameters measured are the same but with different machines.

There are 205k training data samples, 35k for validation, and 20k for testing. The maximum time-series length is 2, with outliers filtered out and short series padded. The number of features is 498, the least among these three data sets. 

The number of measurement tasks is 10, and output dimension is 20. Note that these tasks are not the same as those in P1 and P2 data. The percentage of positive cases is about 1.6\%.

\section{Results} \label{section: results}
The SST models have been run on the three data sets mentioned in the last section. The hyper-parameters are tuned within the range shown in Table. \ref{hyperparameters}, and the best combinations are chosen to present below for each data set. 

To validate the effectiveness of SST, the results are compared with two baseline models. The first one is variance weighted multi-headed quality driven autoencoder (VWMHQAE) \cite{zhang2021autoencoder} which was developed by our team in 2020. The model is based on stacked autoencoder architecture, and utilized the output (quality-control variables) information by reconstructing both the input and output after encoding. It added the multi-headed structure to do the multi-task learning, and applied a variance-based weight to the tasks that are same as SST model as in Equation. \ref{eqn:total_loss}. It has been proven to work well with non-time-series data in our previous experiments with similar sensor data, therefore serves as a good baseline model for SST. Since it doesn't have an architecture to cover the time dimension, the data is flattened before feeding into the model. Also, we trained a second baseline model: a bidirectional LSTM model (Bi-LSTM), which is one of the golden standard models for time series data, to have a comprehensive benchmark on the performance of SST. 

Due to the highly imbalanced nature of the data sets, accuracy would not make much sense to evaluate the models. The most important metrics that the industry cares are True Positive Rate (TPR, also called recall or sensitivity) and False Positive Rate (FPR, also called fall-out or false alarm ratio). However, comparing two metrics together is not intuitive, so we chose to use the Receiver Operating Characteristic (ROC) curve and the Area Under Curve (AUC) as the main metric in this paper. More detailed results are covered in Appendix.

\subsubsection*{P1} 
For the P1 data set, SST model is set as 3 layers, both $dmodel$ and $dff$ are 128, dropout rate is 0.5, batch size is 2048, $n\_heads$ is 1, learning rate factor is 0.5, and the uncertainty based weighting is off. the VWMHQAE model is set as three layers with hidden dimension [512, 256, 128], and Bi-LSTM model with dimension equal to $dff$. All models are followed by a three-layer MLP classifier with all hidden dimensions as $dff$.

\begin{table}[!t]
\renewcommand{\arraystretch}{1.3}
\caption{Result comparison with baseline models: P1}
\label{tbl:P1}
\centering
\begin{tabular}{cccc}
\hline
Task &SST&VWMHQAE&Bi-LSTM\\
\hline
1& $0.70\pm0.12$&$0.63\pm0.12$&$\textbf{0.73}\pm0.13$\\
2& $\textbf{0.60}\pm0.18$&$0.54\pm0.03$&$0.56\pm0.01$\\
3& $\textbf{0.86}\pm0.01$&$0.77\pm0.05$&$\textbf{0.86}\pm0.03$\\
4& $\textbf{0.91}\pm0.01$&$0.89\pm0.01$&$0.88\pm0.01$\\
5& $\textbf{0.55}\pm0.03$&$\textbf{0.55}\pm0.02$&$0.51\pm0.05$\\
6& $0.53\pm0.05$&$\textbf{0.57}\pm0.03$&$0.50\pm0.03$\\
7& $0.64\pm0.02$&$\textbf{0.65}\pm0.01$&$0.64\pm0.01$\\
8& $\textbf{0.82}\pm0.03$&$0.78\pm0.04$&$0.78\pm0.11$\\
9& $0.71\pm0.09$&$\textbf{0.77}\pm0.01$&$\textbf{0.77}\pm0.06$\\
10& $\textbf{0.92}\pm0.03$&$0.82\pm0.03$&$0.67\pm0.23$\\
11& $\textbf{0.89}\pm0.01$&$0.77\pm0.03$&$0.88\pm0.01$\\
\hline
\end{tabular}
\end{table}

\begin{figure}[!t]
\centering
\includegraphics[width=0.9\linewidth]{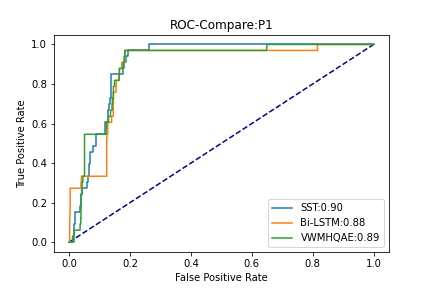}
\caption{ROC curve for SST and baseline models on P1 data set, 4th task}
\label{fig_P1}
\end{figure}

The results for the 11 tasks are summarized in Table. \ref{tbl:P1}. in 7 of the tasks SST are the best performer, especially for the high performing tasks where AUC larger than $0.8$. 

From the results we can also see that some of the tasks have poor results for all three models. They are difficult to get a high AUC value with any model due to the intrinsic complex and noisy nature. Only those measurement tasks with decent results can lead to realistic value in industry applications. This is one of the primary motivations behind our decision to open-access these data sets: researchers all around the world are welcomed to use and explore this data. This will not only help us to gain more understanding about the data sets, but also enrich the research field.

To further illustrate the results, the ROC curve is plotted for the task with highest AUC as in Fig. \ref{fig_P1}. SST has a higher score than the two baseline models, and the curve is smoother, meaning a more even distribution of the prediction probabilities and a finer grid in the prediction space. The source code can be found at https://github.com/Seagate/SoftSensingTransformer. 

\subsubsection*{P2}
SST model is set as 3 layers, both $dmodel$ and $dff$ are 128, dropout rate is 0.3, batch size is 2048, $n\_heads$ is 1, learning rate factor is 0.5, and the uncertainty based weighting is on. Baseline models are the same as P1.

\begin{table}[!t]
\renewcommand{\arraystretch}{1.3}
\caption{Result comparison with baseline models: P2}
\label{tbl:P2}
\centering
\begin{tabular}{cccc}
\hline
Task &SST&VWMHQAE&Bi-LSTM\\
\hline
1& $\textbf{0.89}\pm0.01$&$0.87\pm0.02$&$0.88\pm0.04$\\
2& $0.64\pm0.01$&$\textbf{0.66}\pm0.02$&$0.59\pm0.07$\\
3& $0.60\pm0.06$&$\textbf{0.62}\pm0.01$&$0.55\pm0.01$\\
4& $\textbf{0.85}\pm0.01$&$0.82\pm0.01$&$0.84\pm0.01$\\
5& $0.45\pm0.07$&$\textbf{0.53}\pm0.10$&$0.52\pm0.10$\\
6& $0.64\pm0.02$&$0.71\pm0.04$&$\textbf{0.72}\pm0.02$\\
7& $0.78\pm0.01$&$\textbf{0.81}\pm0.02$&$0.79\pm0.01$\\
8& $\textbf{0.72}\pm0.03$&$0.70\pm0.09$&$0.69\pm0.03$\\
9& $0.71\pm0.12$&$\textbf{0.77}\pm0.08$&$\textbf{0.48}\pm0.14$\\
10& $\textbf{0.76}\pm0.02$&$0.75\pm0.01$&$\textbf{0.76}\pm0.04$\\
11& $0.79\pm0.01$&$0.80\pm0.01$&$\textbf{0.82}\pm0.01$\\
\hline
\end{tabular}
\end{table}

\begin{figure}[!t]
\centering
\includegraphics[width=0.9\linewidth]{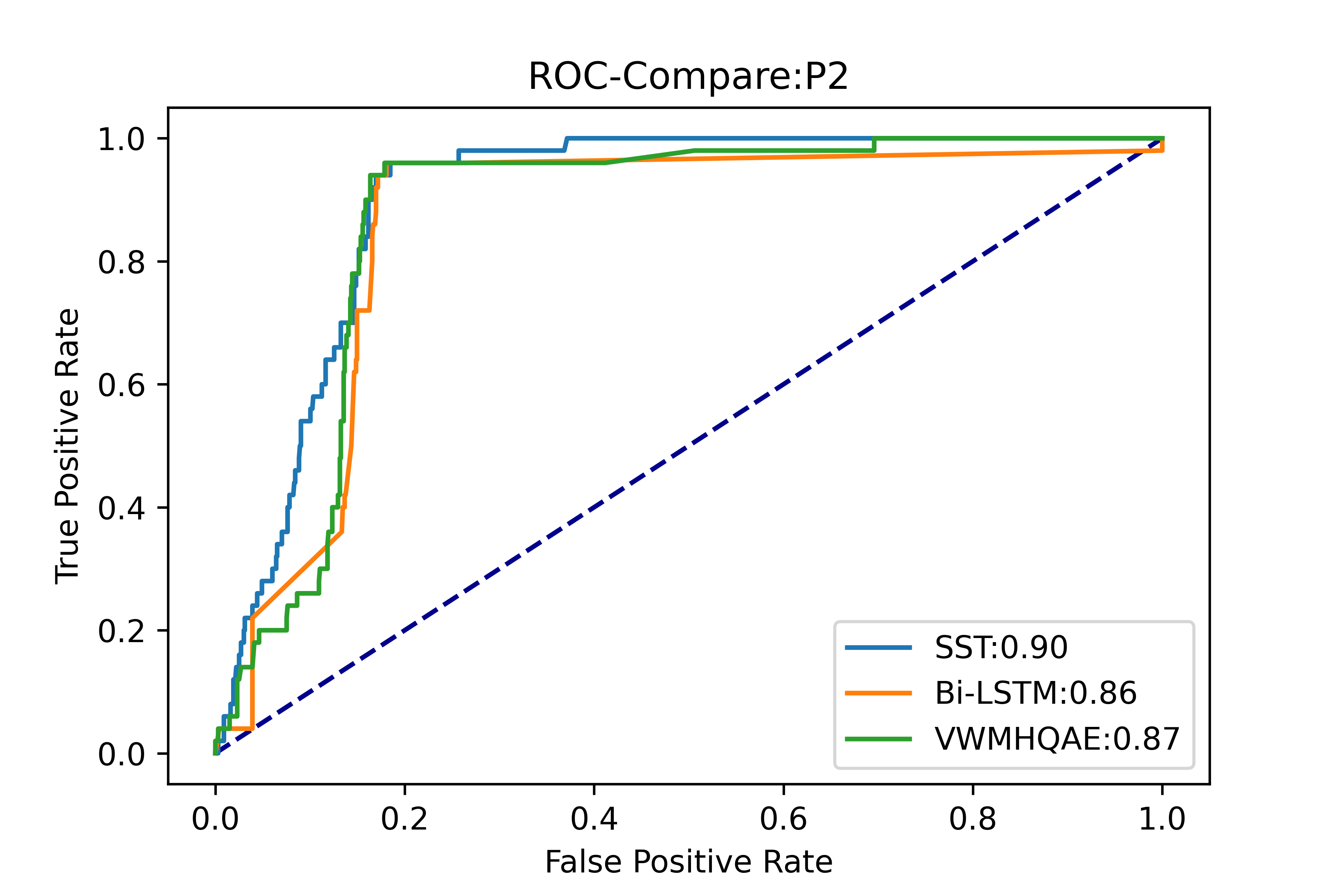}
\caption{ROC curve for SST and baseline models on P2 data set, 1st task}
\label{fig_P2}
\end{figure}

The results for the 11 tasks are summarized in Table. \ref{tbl:P2}. SST is the best performer in 4 of the tasks, including the two tasks with the best prediction. Same as in P1, some of the tasks have poor results for all three models due to the intrinsic complexity and noise in the data set, and we mostly care about the tasks with best results. In this data set, there is only one time step, and as expected the VWMHQAE model, which is not designed for time series data, is showing better results comparing to P1 data, and it has the best performance in 5 out of the 11 tasks.

The ROC curve for the task with highest AUC as in Fig. \ref{fig_P2} is very similar to the previous one. SST is slightly smoother than the baseline models, with a higher AUC. 

\subsubsection*{P3}
SST model is set as 3 layers, both $dmodel$ and $dff$ are 128, dropout rate is 0.3, batch size is 2048, $n\_heads$ is 1, learning rate factor is 0.3, and the uncertainty based weighting is on. Baseline models are the same as P1.

\begin{table}[!t]
\renewcommand{\arraystretch}{1.3}
\caption{Result comparison with baseline models: P3}
\label{tbl:P3}
\centering
\begin{tabular}{cccc}
\hline
Task &SST&VWMHQAE&Bi-LSTM\\
\hline
1& $\textbf{0.42}\pm0.08$&$0.23\pm0.12$&$0.32\pm0.08$\\
2& $\textbf{0.94}\pm0.02$&$\textbf{0.94}\pm0.02$&$0.93\pm0.01$\\
3& $\textbf{0.68}\pm0.08$&$0.53\pm0.10$&$0.56\pm0.05$\\
4& $-$&$-$&$-$\\
5& $0.53\pm0.08$&$\textbf{0.65}\pm0.05$&$0.58\pm0.05$\\
6& $-$&$-$&$-$\\
7& $0.60\pm0.05$&$0.56\pm0.02$&$\textbf{0.61}\pm0.07$\\
8& $\textbf{0.72}\pm0.05$&$0.62\pm0.08$&$0.65\pm0.08$\\
9& $-$&$-$&$-$\\
10& $0.81\pm0.01$&$0.81\pm0.01$&$\textbf{0.82}\pm0.01$\\
\hline
\end{tabular}
\end{table}

\begin{figure}[!t]
\centering
\includegraphics[width=0.9\linewidth]{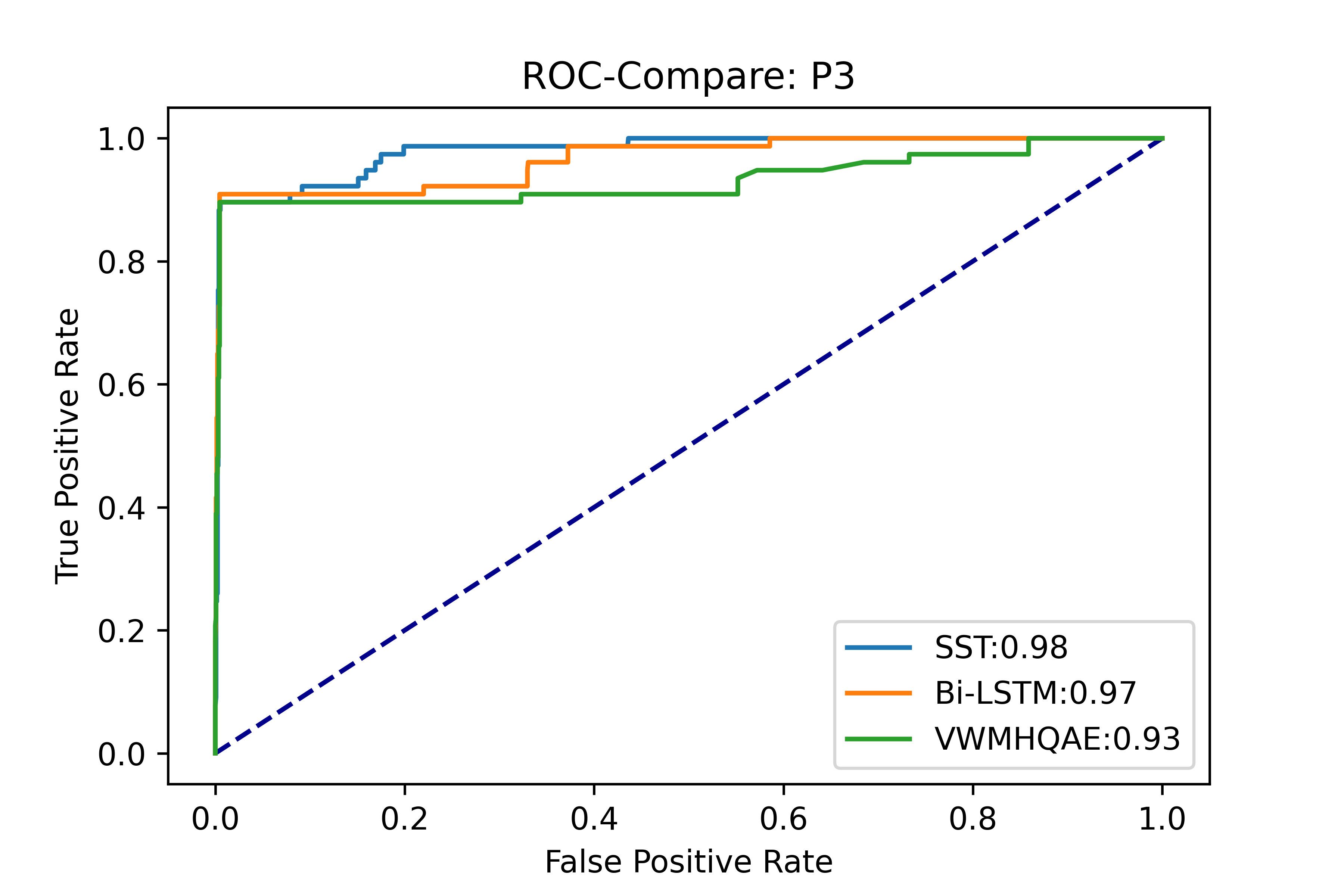}
\caption{ROC curve for SST and baseline models on P3 data set, 2nd task}
\label{fig_P3}
\end{figure}

The results for 7 out of 10 tasks are summarized in Table. \ref{tbl:P3}, because the others has too few testing data samples. SST is the best performer in 4 of the tasks, including the first task with the best prediction. Some of the tasks have poor results for all three models and even with an AUC lower than 0.5, meaning it's worse than a random guess. Its main cause is the distribution shift and further experiments will be carried out when we accumulated more data in Seagate factories. The ROC curve for the task with highest AUC as in Fig. \ref{fig_P3} is very similar to the previous ones.

\section{Discussion and Conclusion} \label{section: conclusion}
We have explored the direct application of Transformers to soft sensing. To our knowledge, we are the first to provide large scale soft sensing data sets, and the first to benchmark the results with the original transformer model in the soft sensing field. Also, this is the first time that transformer model goes beyond human in the sense that the input data is not human-interpretable. We analogize the time-series data as a sequence of sensor values, each time step is taken as a word, and process the sentence-like data by a standard transformer encoder exactly as in NLP. This direct and intuitive strategy has shown an exciting result for our data sets that outperforms our previous model and Bi-LSTM model.

We share these data sets with the excitement of advancing interest and work in research and applications of soft sensing. We invite future work into the exploration of improving SST performance on some tasks that have been particularly challenging in our experiments to learn. Another future direction can be the examination of appropriate time sequences for these data sets, and exploration of better ways to address missing data. We are working on acquiring more data with longer sequences, to better understand the impact of time series length in the prediction of quality. In the meantime, we have provided three data sets to cover a variety of sensors, and to examine the generalizability of deep learning models, and we believe these data sets can enrich the soft sensing research field and serve as one of the standard tools to evaluate the effectiveness of future research.


\ifCLASSOPTIONcompsoc
  \section*{Acknowledgments}
\else
  \section*{Acknowledgment}
\fi

The authors would like to thank Seagate Technology for the support on this study, the Seagate Lyve Cloud team for providing the data infrastructure, and the Seagate Open Source Program Office for open sourcing the data sets and the code.  Special thanks to the Seagate Data Analytics and Reporting Systems team for inspiring the discussions.



\bibliographystyle{IEEEtran}
\bibliography{ref}
%



\newpage
\section*{Appendix}

\begin{figure}[!ht]
\centering
\includegraphics[width=0.9\linewidth]{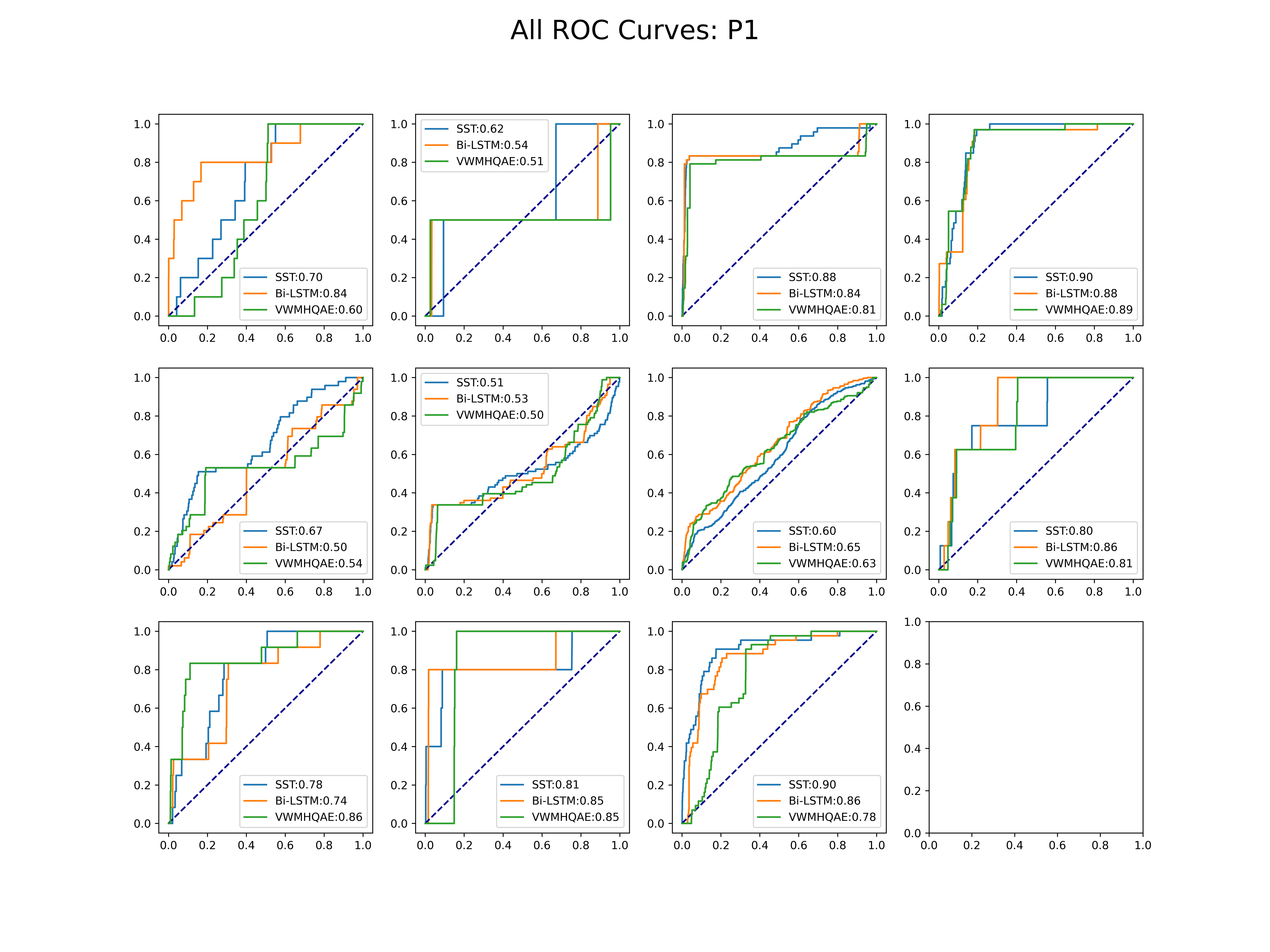}
\caption{ROC curves for SST and baseline models on P1 data set, all tasks}
\end{figure}

\begin{figure}[!ht]
\centering
\includegraphics[width=0.9\linewidth]{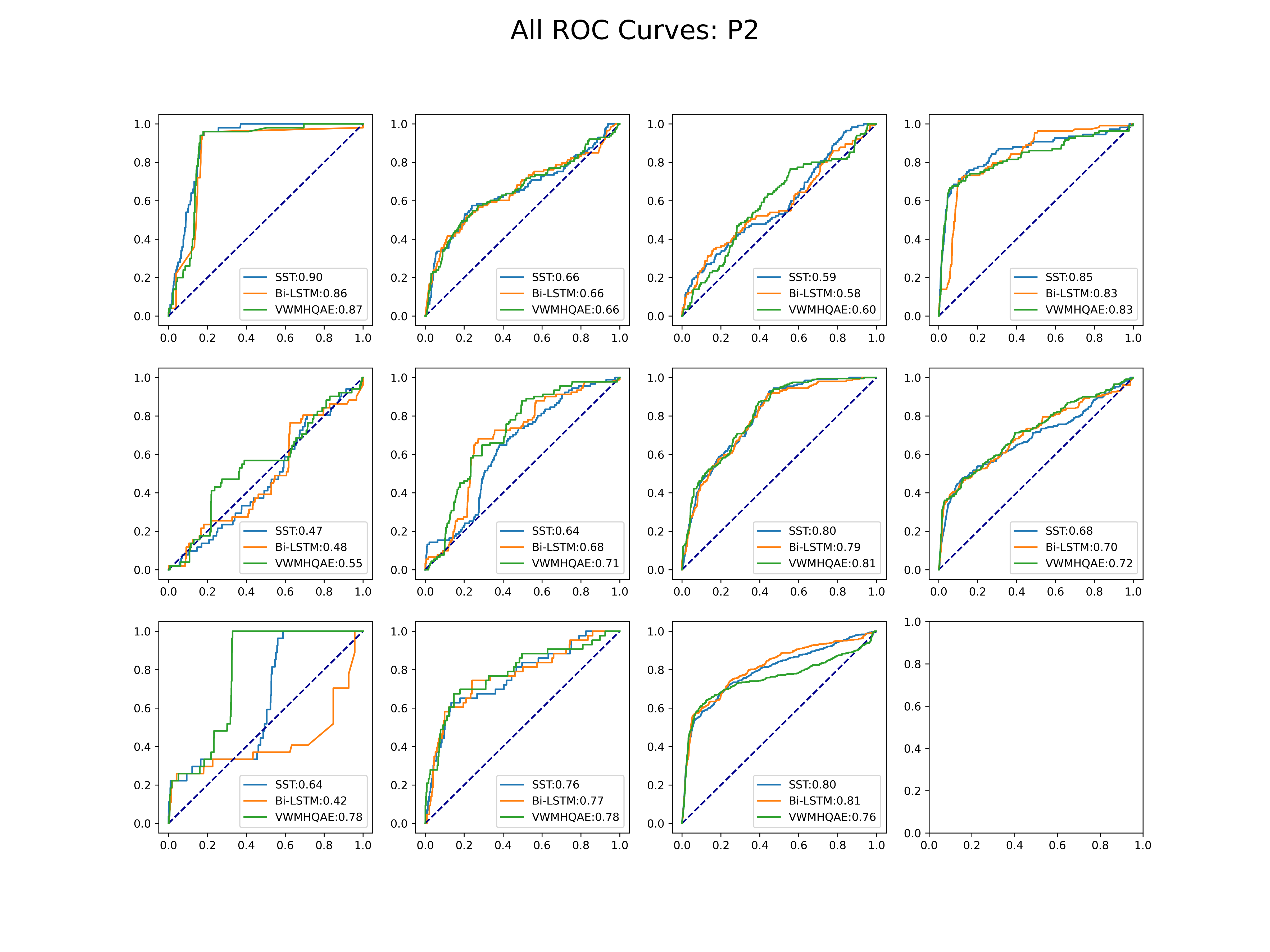}
\caption{ROC curves for SST and baseline models on P2 data set, all tasks}
\end{figure}

\begin{figure}[!ht]
\centering
\includegraphics[width=0.9\linewidth]{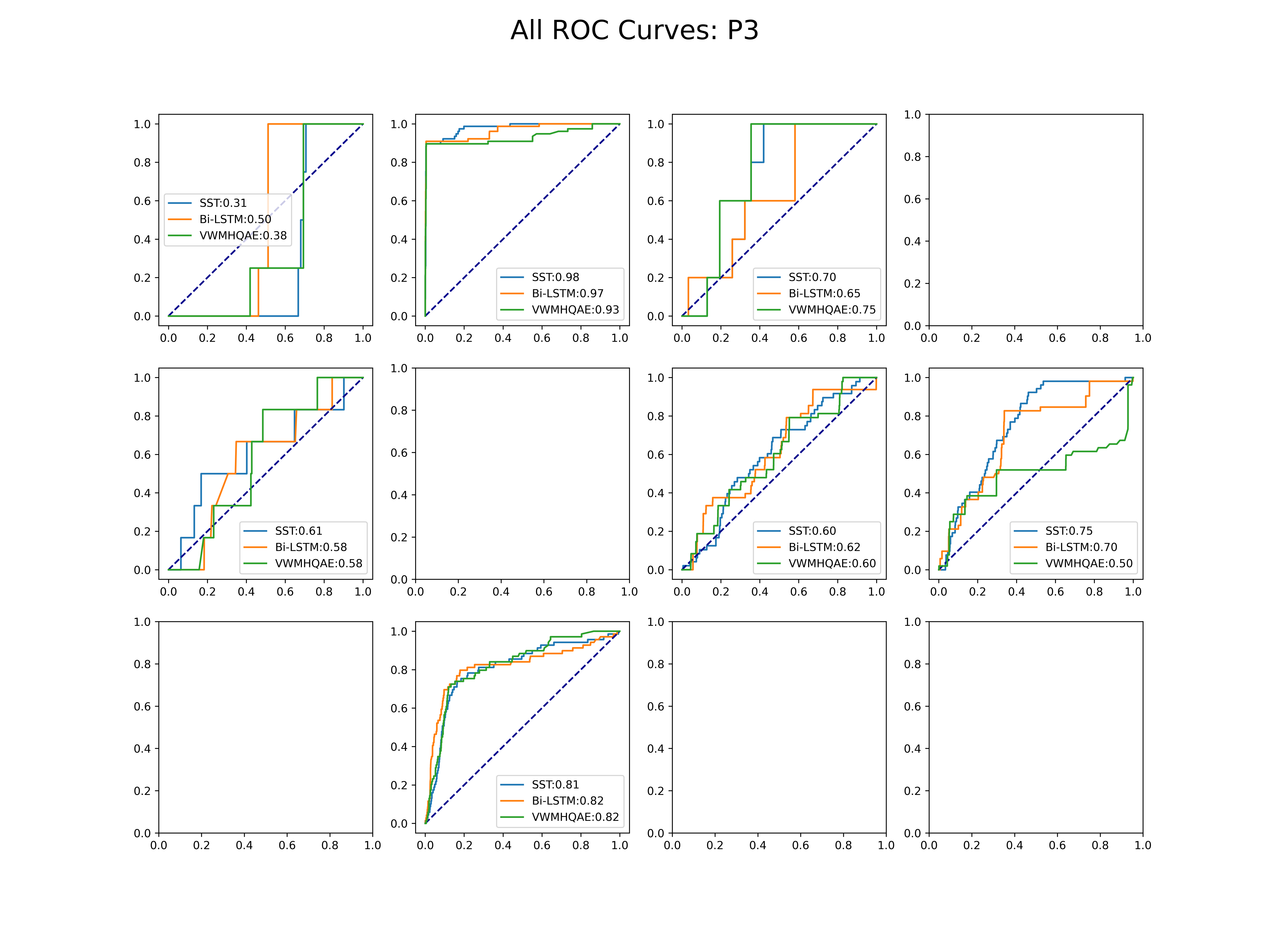}
\caption{ROC curves for SST and baseline models on P3 data set, all tasks}
\end{figure}

\begin{figure}
\centering
\includegraphics[width=\linewidth]{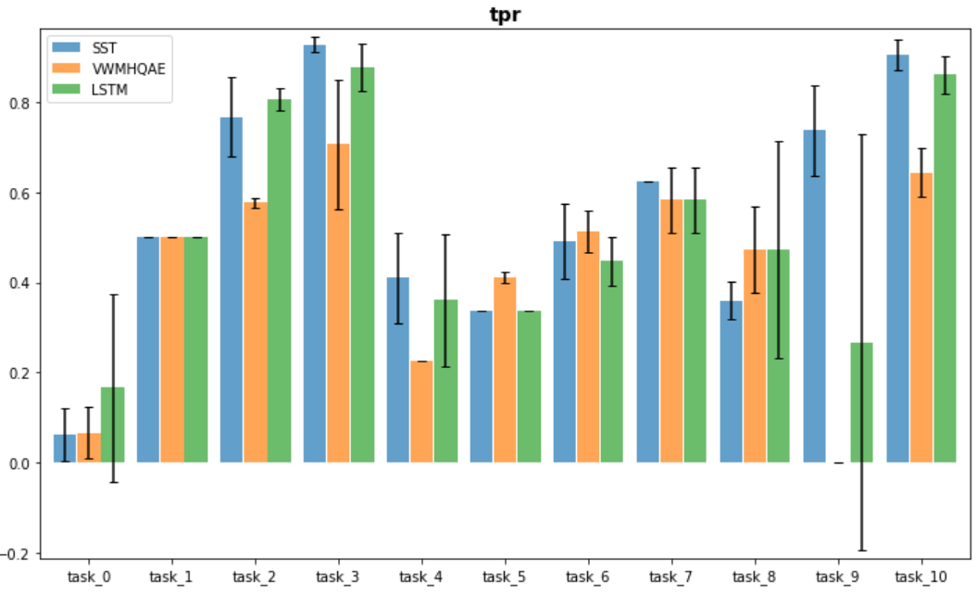}
\caption{TPR for SST and baseline models on P1 data set}
\end{figure}

\begin{figure}
\centering
\includegraphics[width=\linewidth]{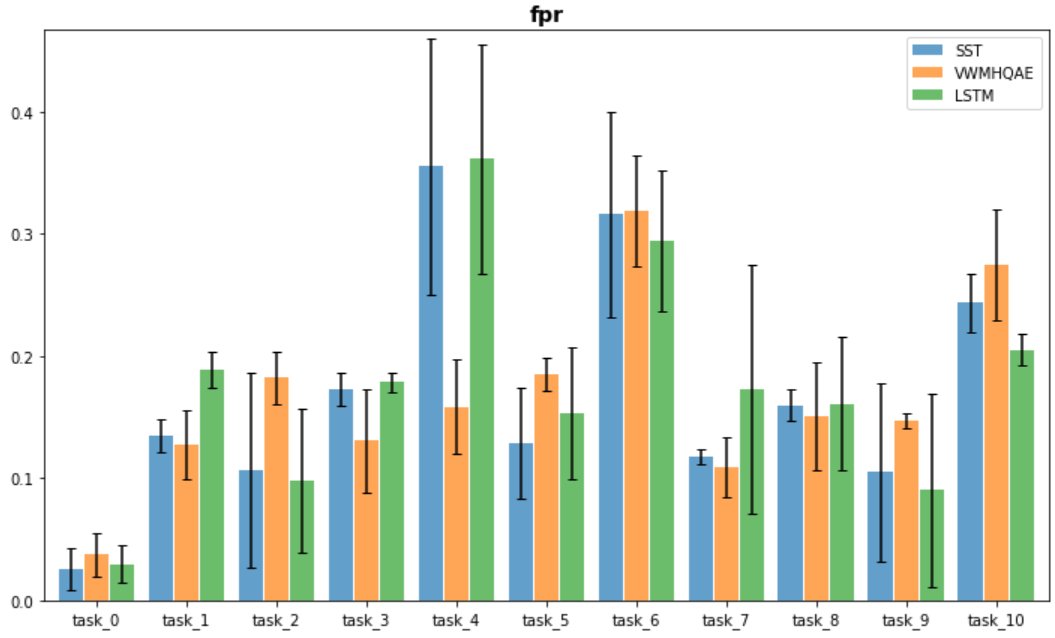}
\caption{FPR for SST and baseline models on P1 data set}
\end{figure}

\begin{figure}
\centering
\includegraphics[width=\linewidth]{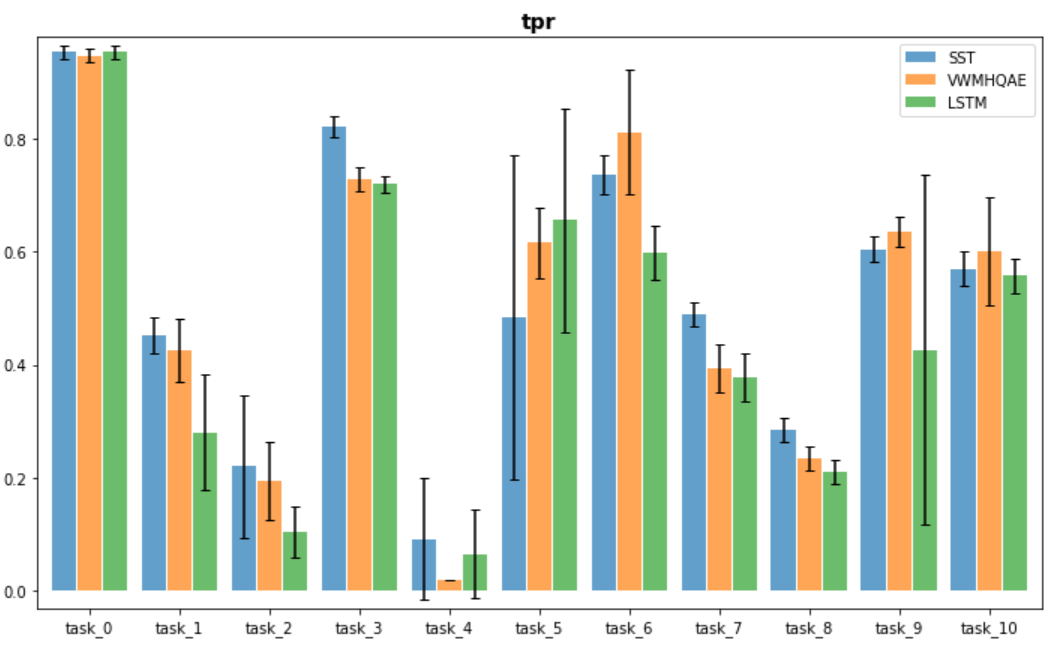}
\caption{TPR for SST and baseline models on P2 data set}
\end{figure}

\begin{figure}
\centering
\includegraphics[width=\linewidth]{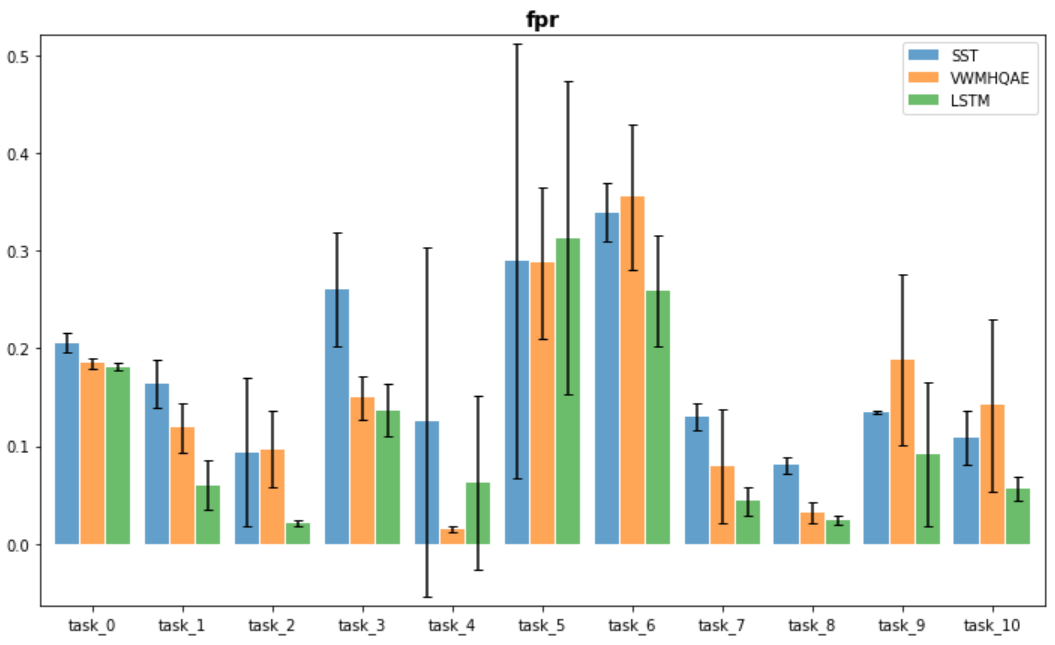}
\caption{FPR for SST and baseline models on P2 data set}
\end{figure}

\begin{figure}
\centering
\includegraphics[width=\linewidth]{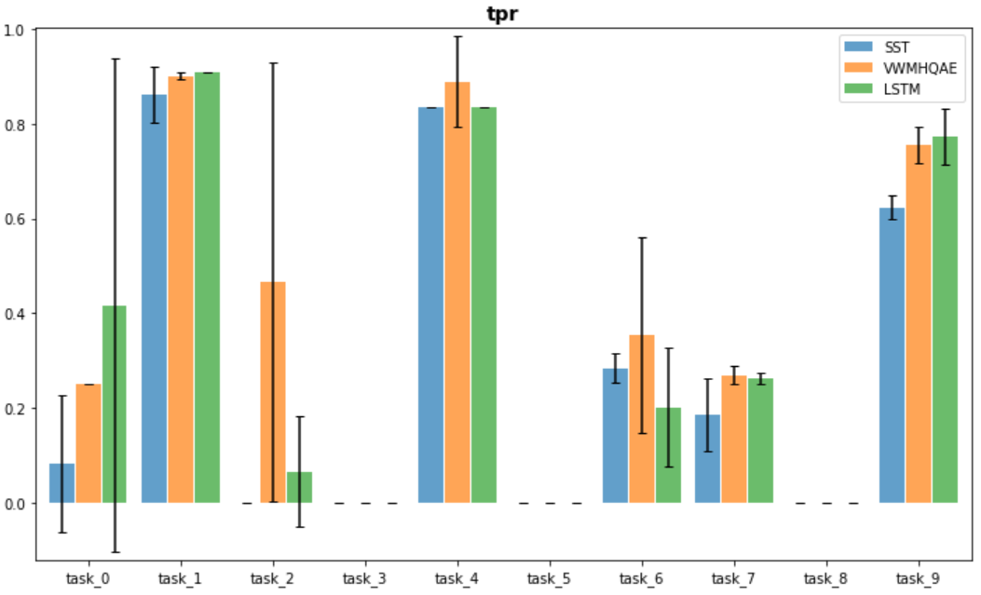}
\caption{TPR for SST and baseline models on P3 data set}
\end{figure}

\begin{figure}
\centering
\includegraphics[width=\linewidth]{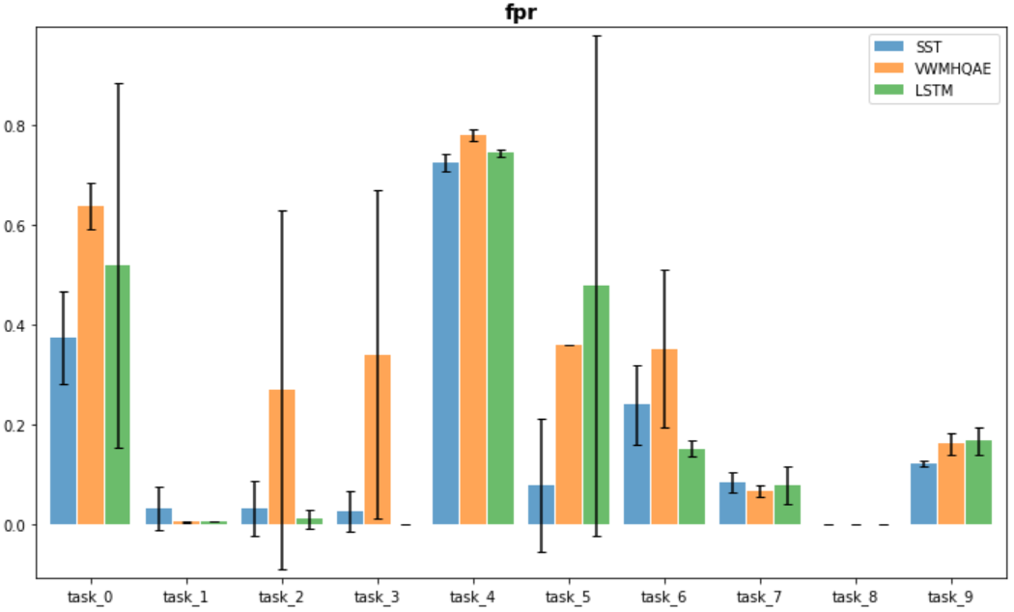}
\caption{FPR for SST and baseline models on P3 data set}
\end{figure}

\end{document}